\documentclass[11pt,a4paper]{article}

\usepackage[T1]{fontenc}
\usepackage[utf8]{inputenc}
\usepackage[english]{babel}
\usepackage{underscore} 
\usepackage{lineno} 

\usepackage{csquotes}
\usepackage{amsmath}
\usepackage{amsfonts}
\usepackage{amssymb}
\usepackage{xcolor,lipsum}
\usepackage{booktabs}
\usepackage{fixltx2e}
\usepackage{setspace}
\usepackage{threeparttable}
\usepackage{graphicx}
\usepackage{wrapfig}
\usepackage{epstopdf}
\usepackage{lscape}
\usepackage{rotating}
\usepackage{pdflscape}	
\usepackage{subcaption}  
\usepackage{longtable}
\usepackage{fancyvrb}
\usepackage{url} 
\usepackage{enumerate}
\usepackage[colorlinks=true,linkcolor=blue,citecolor=blue,urlcolor=blue,breaklinks]{hyperref}

\usepackage{listings} 
\usepackage{color}  
\usepackage{chngcntr}

\definecolor{dkgreen}{rgb}{0,0.6,0}
\definecolor{gray}{rgb}{0.5,0.5,0.5}
\definecolor{mauve}{rgb}{0.58,0,0.82}
\definecolor{darkblue}{rgb}{0.055,0.094,0.588}
\definecolor{darkred}{rgb}{0.4,0,0.0157}
\definecolor{myblue}{rgb}{0.2,0.2,0.7}
\definecolor{myred}{rgb}{0.9,0,0}

\usepackage[style=authoryear,autocite=inline,backend=biber,sorting=nyt,natbib=true]{biblatex}
\begin{document}


\author{~\\
	\textsc{Artur Tarassow} \\
	{\small {atecon@posteo.de} }
  }
\title{The potential of LLMs for coding with low-resource and domain-specific programming languages
	\thanks{%
		I am grateful for the helpful comments of Steffen Remus, Allin Cottrel, Riccardo (Jack) Luchhetti and Sven Schreiber, and for the discussion of participants at the 8th gretl conference in Gdańsk (June 2023). Any errors or omissions are the sole responsibility of the authors.
		}
}

\date{\today}

\maketitle

\begin{abstract}
	This paper presents a study on the feasibility of using large language models (LLM) for coding with low-resource and domain-specific programming languages that typically lack the amount of data required for effective LLM processing techniques. This study focuses on the econometric scripting language named hansl of the open-source software gretl and employs a proprietary LLM based on GPT-3.5. Our findings suggest that LLMs can be a useful tool for writing, understanding, improving, and documenting gretl code, which includes generating descriptive docstrings for functions and providing precise explanations for abstract and poorly documented econometric code. While the LLM showcased promoting docstring-to-code translation capability, we also identify some limitations, such as its inability to improve certain sections of code and to write accurate unit tests. This study is a step towards leveraging the power of LLMs to facilitate software development in low-resource programming languages and ultimately to lower barriers to entry for their adoption.
\end{abstract}



\newpage


\newpage
\section{Introduction}
\label{sec:intro}

Large language models (LLMs) have emerged in computer science and are revolutionizing the interplay of human (natural) language and computation. The most well-known ones are GPT-3.5 (a base LLM) and ChatGPT\footnote{\url{https://chat.openai.com/}} (an instruction-tuned LLM), both of which have gained significant attention for their impressive language generation capabilities.\footnote{You may think that "revolutionizing" may be exaggerated. However, researchers from Microsoft Research argue that the recent GPT-4 model "[...] could reasonably be viewed as an early (yet still incomplete) version of an artificial general intelligence (AGI) system" \parencite{Bubeck.etal2023}.}

Automatic code generation is a long-standing challenge in computer science \parencite{Manna.Waldinger1971}. Large corpora comprising code and the training of huge language models have fuelled research in this direction. The potential benefits of using LLMs for programming comprise assistance with code creation, documentation, explanation, and description. For instance, LLMs may assist with naming variables and functions, help with code refactoring.

Previous studies have investigated the effectiveness of LLMs as a programming assistant for various tasks typically faced by programmers. However, most of the research has focused on high-resource general-purpose programming languages, such as Python, C++, or JavaScript \parencite{Chen.etal2021, Wang.Komatsuzaki2021, Tian.etal2023}. There has been only limited research on low-resource programming languages, which often lack large amounts of data, such as source code or references, making them challenging for LLMs to process compared to high-resource languages. Recent exceptions are \textcite{Zuegner.etal2021, Ahmed.Devanbu2022a, Chen.etal2022} who studied the performance of LLMs for coding with low-resource but general-purpose programming languages.

Domain-specific programming languages also exist and offer high levels of abstraction. Due this they can be more efficient than general-purpose languages for specific domain-related tasks, such as financial modelling, econometrics, or scientific computing. However, current research on LLMs as programming assistants has focused only on a limited subset of programming languages, leaving the vast majority understudied. The aim of this paper is to investigate how LLMs may assist with the use of a low-resource \textit{and} domain-specific programming language. By exploring the potential of LLMs in this context, this paper aims to contribute to filling the gap in research on such languages.

For this study, we choose scripting language named hansl. Hansl (a recursive acronym: “hansl's a neat scripting language”) is the scripting language of gretl\footnote{\url{gretl.sourceforge.net}}, an open source econometrics and statistics package written in C and licensed under the GNU GPL \parencite{Cottrell2017}.\footnote{For a practical introduction to gretl, see e.g. \textcite{Tarassow2019}.} It can compete with the leading paid econometrics packages of Stata and Eviews, as well as the top open-source statistical software project, R.\footnote{\url{https://www.r-project.org/}} Gretl is mainly used by economists and econometricians when doing applied work as well as simulations.\footnote{Note, in order to avoid confusion, we use gretl as a synonym for hansl throughout the text.}

While Python and R code are highly prevalent, the amount of gretl code available is relatively limited.\footnote{We did a search on \url{github.com} to identify files relevant to gretl and hansl. The search query used was as follows: \texttt{("gretl" OR "hansl") AND (path:*.inp OR path:*.gfn)}. This query searched for files that contained either the term "hansl" or "gretl" within files with either the extension \texttt{inp} (gretl script file) or \texttt{gfn} (gretl package file). The search results showed that there were 1500 such files available on GitHub. It is notable that the number of files available for R and Python are multiple factors higher in comparison: Searching for \texttt{language:R} revealed a total of 5.1 million files while searching for Python on GitHub revealed ten times as many files. The search was conducted on the 2023-07-18.}
This means that the data available for training LLMs to perform specific tasks using gretl code is far less comprehensive than that available for Python or R code. As a result, existing LLMs may not be as effective when applied to gretl code tasks used in applied research. Despite these limitations, we demonstrate that LLMs can still be applied successfully to certain coding tasks using gretl. Specifically, in the field of (applied) econometrics, LLMs can help users understand user-written functions involving linear algebra and statistics. They can also provide suggestions for coding solutions based on natural language prompts. Overall, LLMs have the potential to save time, minimize or avoid coding errors, and increase productivity for both programmers, econometricians, and other knowledge workers. This indicates that publicly existing models are able to generalize well to low-resource programming languages \parencite{Zuegner.etal2021}.

We conduct a series of tasks applied coding economists or econometricians may face such as code documentation, code explanation and description (with special focus on the econometrics involved), variable and function naming, code creation as well as refactoring of code. We also present examples on how gretl code can be written by the LLM based on a docstring only. We show that the LLM also writes gretl code for computing the Fibonacci sequence or the root mean squared error including only minor syntactical errors. In a last experiment, we simulate a simple coursework exercise for an introductory econometric class and show how the LLM writes the gretl script according to the class instructions to some extent.
For this, we use a proprietary instruction-based LLM provided for free (after registration) by \url{www.you.com}. The model is a generative pre-trained transformer using the ChatGPT framework which has around 175 billion parameters and is based on GPT-3.5.

It should be said that we were unable to use a "vanilla" pre-trained model. The publicly usable model relies on human-in-the-loop (HITL) feedback to improve its performance.\footnote{HITL is a machine learning method that involves incorporating human feedback at various stages of training in order to enhance the accuracy and overall model performance. This is particularly important in the context of natural language processing, where the intricacies of language can be difficult for models to interpret without additional guidance.}
Since the beginning of June 2023, the author has actively used the model and provided feedback on gretl code-related prompts, allowing the model's performance to continually improve over time. However, this introduces a potential challenge in assessing the model's true capabilities, as it becomes difficult to determine whether the model can generate working gretl code on its own, or if it relies solely on the HITL feedback to do so. While this issue is not the focus of our paper, we recognize the potential impact it may have on the results.

In the next section, we provide a brief background on LLMs and summarize the recent literature on both the use of LLMs for programming generally as well as on low-resource programming languages specifically. Afterwards, we present the experiments of applying the LLM to gretl code and discuss the results. The paper ends with a conclusion.

\section{Background}

\subsection{Large Language Models}
Modern LLMs rely on deep learning algorithms and large datasets to process and understand human language and generate text. Nowadays, they use the transformer architecture, which were introduced by Google in 2017, as the backbone of their models \parencite{Vaswani.etal2017}. The transformer architecture is a neural network that is capable of processing long sequences of input data in parallel, resulting in more efficient and effective training.

The key innovation of the transformer is the self-attention mechanism, which allows the model to identify which parts of the input are relevant for each output. This enables the model to identify and learn more complex relationships between the input and output, resulting in better performance on a wide range of natural language processing tasks. For details see also \textcite{Radford.etal2019, Wolfram2023}. Current architectures can be autoregressive models that have a memory of the last $N$ word-pieces (potentially solely parts of a word). Their pre-training objective is to predict the next token and the next sentence.

LLMs are used for many tasks such as machine translation, sentiment analysis, text completion, summarization and extraction etc.

\subsection{Large language models used for programming}

Automatic code generation is a longstanding challenge \parencite{Manna.Waldinger1971}. Large corpus comprising code \parencite{Husain.etal2020} and the training of huge language models such as GPT-J \parencite{Wang.Komatsuzaki2021} have fuelled research in this direction.\footnote{GPT-J is a large-scale language model that has been trained on a massive amount of text data and is general in nature, meaning it can perform a wide range of tasks such as text generation, translation and summarization. While GPT-J is powerful enough to generate code, it is not specifically tailored for this task like the well-known "code-davinci" models are.} In the following, we provide a brief overview of recent research on the use and effectiveness of LLMs for programming.

\textcite{Chen.etal2021} introduce Codex, a GPT language model fine-tuned on publicly available code from GitHub, and evaluate its Python code-writing capabilities. The study reveals that Codex significantly outperforms GPT-3 and GPT-J on a new evaluation set designed to measure functional correctness for synthesizing programs from docstrings.\footnote{A docstring is a string literal specified in the source code that is used to document a specific segment of code such as a function. It is used to provide information on what a particular piece of code does, its input parameters, expected return values, and any relevant usage example}

The company GitHub published its product Copilot which relies on OpenAI's Codex model.\footnote{\url{https://github.com/features/copilot/}} Copilot is a programming assistant for various general-purpose programming languages widely used in industry. The model is trained on billions lines of code to turn natural language prompts into coding suggestions.

\textcite{Xu.etal2022} find that existing open-source models perform similar to closed-source models such as Codex in some programming languages, although targeted mainly for natural language modelling.
The authors publish a LLM named PolyCoder which is the first public model trained exclusively on code from multiple programming languages. Also, the authors claim that training on natural language text and code jointly can benefit code modelling.

Recently, \textcite{Tian.etal2023} studied the potential of ChatGPT as a programming assistant on code generation, detecting and fixing bugs as well as code summarization. They compare the model against two benchmark approaches using Python as the programming language being evaluated. One benchmark is using LeetCode which is an online platform offering a diverse range of programming problems for software engineering interviews, where new problems are constantly posted and updated. The second benchmark is named Refactory which is a semantic-based assignments repair tool written in Python and released in 2019. ChatGPT dominates the benchmarks on code writing and performs equally well to Refactory on program repair. On code summarization their results show that ChatGPT may have problems in explaining the intention of code though.

\textcite{Yuan.etal2023} evaluate the performance of ChatGPT for writing automatic unit tests. ChatGPT's generated tests were analyzed to evaluate its correctness, sufficiency, readability, and usability. The results showed that while the tests still suffer from correctness issues, passing tests generated by ChatGPT resemble manually-written tests in terms of coverage, readability, and developer preference, suggesting that generating unit tests with ChatGPT could be promising.

\textcite{Bubeck.etal2023} evaluate GPT-4 for coding challenges. They find that GPT-4 does much better on the HumanEval, a docstring-to-code dataset consisting coding problems, compared to state-of-the-art LLMs trained specifically on code. The authors also compare leading LLMs on the LeetCode dataset and state that "GPT-4 significantly outperforms the other models, and is comparable to human performance..." \parencite[ch. 3.1.1]{Bubeck.etal2023}. They even go further and find that GPT-4 can be used to create complex data visualization, write a 3D game in HTML with JavaScript, using a very high-level specification, customize an deep-learning optimizer module involving operations such as applying SVD and linear algebra just by giving human language description. Furthermore, GPT-4 reverse-engineers assembly code, reasons about code execution of C code and runs a complex Python method by simulating the steps instead of running the program actually.

\textcite{Korinek2023} is a rare exception studying the value-added of LLMs for economic research including coding. Apart from writing Python code for computing and plotting the Fibonacci sequence, he shares his experience that Open AI's model name \texttt{text-davinci-003} is not yet able to write code for simulating basic economic problems like optimal consumption smoothing. According to Korinek, current LLMs can also be used to translate short pieces of Python into Matlab code or debug Python functions. For more complex code human assistance is still required, though.

\subsection{Low-resource programming language and coding}
Low-resource programming languages (LRPLs) are characterized by a low number of records in their datasets, which means that there is little information, or even none, comprising source code or references. This lack of information makes it challenging for language models to process LRPL data. Although models can be trained to achieve high accuracy for tasks such as intent classification and sequence labelling, these tasks require large amounts of labelled data in the target language.

However, due to their scarcity of resources, LRPLs are less studied. This limitation is similar to that experienced in natural language processing of low-resource languages, as elucidated by \textcite{Mastel.etal2023}. The authors study the limited availability of natural language understanding resources for African languages, most of which are considered low resource languages.

Despite the importance of LRPLs in software development, little research exists on how pre-trained LLMs perform on them. Exceptions are \textcite{Zuegner.etal2021} who have demonstrated that training a model with data from multiple programming languages enhances its performance on individual languages. This finding is especially true for LRPLs and suggests that multilingual LLMs may be able to generalize well across programming languages. Recent research by \textcite{Ahmed.Devanbu2022a} also supports this view, particularly in the areas of code summarization, code retrieval, and function naming.

On the other hand, \textcite{Chen.etal2022} focus on the low-resource (general-purpose) programming language Ruby. The authors fine-tune both monolingual and multilingual pre-trained language models for this language, respectively, and evaluate their performance in code search and code summarization. Their results indicate that multilingual models perform less well than monolingual ones.

In their recent work, \textcite{Gong.etal2022} have introduced the MultiCoder LLM, which aims to improve code completion performance on LRPLs. The authors achieve this by adding new layers to a pre-existing LLM in order to enhance its performance, demonstrating that MultiCoder outperforms their monolingual baselines.

For our study, it is relevant to examine the generalization ability of popular LLMs such as GPT-3.5 to the programming language used in the open-source software gretl. While existing research focuses on low-resource but \textit{general} programming languages, we put our focus on a \textit{domain-specific} LRPL.

\section{Tasks}
This section explores several programming tasks that are relevant to applied economists and econometricians. Firstly, we discuss how to improve the documentation of gretl code by writing docstrings based on the code itself. In the following section, we demonstrate how the LLM can be used to translate docstrings into executable code. Moving on, we show how the LLM can help summarize or describe code, and suggest self-explanatory variable and function names for better readability and maintainability of the code. We then further explore how the LLM can aid in refactoring gretl code and writing unit tests. Finally, we provide a brief example of how the LLM can be utilized in translating real-world student exercises from human language into gretl code.\footnote{All exercises were conducted in June 2023. Since mid of July 2023, \url{www.you.com} has switched from GPT-3.5 to GPT-4.0.}

\subsection{Create docstrings from code bodies}
\label{sec:code_to:doc}

In the software industry, the lack of high-quality code documentation is a common challenge. Docstrings are commonly used in code to address this issue. Docstrings provide clear and concise explanations of the code's functionality, simplifying collaboration among developers. Docstrings also help ensure consistency of documentation across different code bases and modules. Additionally, API documentation can be generated automatically from docstrings. Moreover, docstrings can be used to create unit tests, which can improve code quality over time. However, creating and updating docstrings is labor-intensive, which has prompted research into automatic code summarization \parencite{Sun.etal2023}. This paper investigates whether it is feasible to generate docstrings from gretl code bodies following an exercise by \textcite{Chen.etal2021}.

We examine the creation of a docstring for the gretl function \texttt{add_prefix_to_any} in Listing \ref{code:p_write_docstring_add_prefix_to_any}. The function accepts an array of strings as input and adds a prefix followed by each array item, returning the updated array. In case the input array \texttt{targets} is empty, the function throws an error. To avoid confusion, we have enclosed the relevant code in a block of triple backticks.

\begin{lstlisting}[caption={Prompt to write a docstring for \texttt{add_prefix_to_any()}}, label={code:p_write_docstring_add_prefix_to_any}]
Write a docstring for a gretl function. The gretl function is provided in triple backtics:
```
function strings add_prefix_to_any (const strings targets, const string prefix)
	strings ret = array(nelem(targets))
	errorif(nelem(ret) == 0, "Array is empty.")

	if nelem(prefix) == 0
		return targets
	endif

	loop foreach i targets
		string ret[i] = sprintf("%s%s", prefix, "$i")
	endloop

	return ret
end function
```
\end{lstlisting}

The resulting docstring for the \texttt{add_prefix_to_any} function is provided in Listing \ref{code:a_write_docstring_add_prefix_to_any}. The description accurately conveys the purpose of the function, while the input parameter descriptions and types are precise. Moreover, the docstring thoroughly explains the function's output. Remarkably, the LLM recognizes the error case, even though gretl does not support exceptions (specifically, of type "ValueError").

\begin{lstlisting}[caption={Answer on writing a docstring for \texttt{add_prefix_to_any()}}, label={code:a_write_docstring_add_prefix_to_any}]
Add a prefix to each element in an array of strings.

:param targets: The array of strings to modify.
:type targets: array of string
:param prefix: The prefix to add to each element in the array.
:type prefix: string
:return: A new array of strings with the prefix added to each element.
:rtype: array of string
:raises: ValueError if the input array is empty.
\end{lstlisting}

The first example may appear simple, and it is not surprising that the LLM can accurately generate a docstring. However, it is important to note that (1) the LLM is most likely not explicitly trained on gretl's scripting language, and (2) the datatype \texttt{strings} is not present in other popular programming languages.

The next example involves a much more complex function called \texttt{kernel_loop}, taken from the "ketvals" package (v0.91) by  \textcite{Lucchetti.Valentini2023}. The function name and the parameter names passed are not very meaningful, as they consist mostly of single letters (e.g. \texttt{y} or \texttt{Z}), and there are no comments. Additionally, the function calls another private function called \texttt{kernelmat}, which the LLM has not seen before. In order to write an accurate and precise docstring, deep domain knowledge is required, as there are mathematical operations (involving linear algebra) and a loop block. The prompt for writing a docstring can be found in Listing \ref{code:p_docstring_kernel_loop}.

\begin{lstlisting}[caption={Prompt to write a docstring for \texttt{kernel_loop()}}, label={code:p_docstring_kernel_loop}]
Write a docstring for a gretl function. The gretl function is provided in triple backtics:
```
function matrix kernel_loop (const matrix y,
								const matrix X,
								const matrix Z,
								scalar h,
								scalar type[0:4:1],
								matrix optparm[null])
	scalar q = cols(y)
	scalar k = cols(X)
	scalar T = rows(y)
	matrix ret = mshape(NA,T,q*k)
	W = kernelmat(T, type, h, optparm)
	ini = T + 1
	fin = 2*T
	loop t = 1 .. T
		matrix PX = W[ini:fin] .* Z
		matrix num = PX'y
		matrix den = PX'X
		ret[t,] = vec(den\num)'
		ini--
		fin--
	endloop
	return ret
end function
```
\end{lstlisting}

The output generated by the LLM for the \texttt{kernel_loop} function is displayed in Listing \ref{code:a_docstring_kernel_loop}. The docstring correctly identifies the function as executing a kernel regression. Surprisingly, despite the non-meaningful and single-letter naming of the parameters, the LLM still manages to accurately describe them. For example, the LLM refers to the \texttt{h} parameter as the kernel bandwidth parameter, which is correct.

There is only one error (coloured in red) in the LLM's output: for the \texttt{type} parameter, the LLM incorrectly states that it is an array of integers, when in fact it is a scalar value. Although an array of integers is not supported by gretl, the syntax \texttt{scalar type[0:4:1]} used in the function signature indicates the minimum, maximum, and default values, rather than an array of integers.

Interestingly, the LLM identifies an error in the function even though no explicit error is raised in the code body. This is because the LLM recognizes that a linear algebra operation such as matrix multiplication can throw an error if the dimensions of the matrices do not match.

\begin{lstlisting}[caption={Answer on writing a docstring for \texttt{kernel_loop()}. Syntactical error coloured in red.}, label={code:a_docstring_kernel_loop}]
Perform a kernel regression using a loop and return the parameter estimates.

:param y: The dependent variable matrix.
:type y: matrix
:param X: The independent variable matrix.
:type X: matrix
:param Z: The kernel weight matrix.
:type Z: matrix
:param h: The kernel bandwidth parameter.
:type h: scalar
:param type: An array of integer values representing the kernel type.
:type type: @\color{red}array of int@
:param optparm: Optional parameters used to calculate the kernel.
:type optparm: matrix or None
:return: A matrix containing the parameter estimates for kernel regression.
:rtype: matrix
:raises ValueError: If the input arrays are not correctly formatted.
\end{lstlisting}

\subsection{Create executable code from docstrings}
Docstrings provide a clear and concise description of the code's functionality and purpose. By translating these descriptions into actual code, programmers can more easily understand the intended behavior of the program and ensure that they are implementing the desired functionality. Additionally, this process can help improve the accessibility of code to non-technical stakeholders or users who may not have a programming background but want to understand the software's functionality. Thus, translating docstrings into programming code is relevant, necessary, and contributes significantly to the quality and usability of software applications.

\textcite{Chen.etal2021} have recently found evidence that current-state LLMs are able to write executable Python code based on information contained in docstrings only. \textcite{Bubeck.etal2023} evaluate the ability of the GPT-4 model to translate docstrings to code and show that this LLM is comparable to human levels.

In this exercise, we want the LLM to write executable gretl code based on a docstring. As an initial example, we make use of the docstring the LLM provided in Subsection \ref{sec:code_to:doc} for function \texttt{add_prefix_to_any()}. We start by opening a new session to clear the LLM's memory. Listing \ref{code:p_write_add_prefix_to_any} presents the prompt for creating a gretl function.

\begin{lstlisting}[caption={Prompt to write a gretl function based on a docstring}, label={code:p_write_add_prefix_to_any}]
Write a gretl function based on the following docstring. The docstring is in triple backtics.
```
Add a prefix to each element in an array of strings.

:param targets: The array of strings to modify.
:type targets: array of string
:param prefix: The prefix to add to each element in the array.
:type prefix: string
:return: A new array of strings with the prefix added to each element.
:rtype: array of string
:raises: ValueError if the input array is empty.
```
\end{lstlisting}

Listing \ref{code:a_write_add_prefix_to_any} presents the proposed function and a detailed description. While the LLM provides a code skeleton that serves as a good starting point, the code itself contains several syntax errors. For instance, the function signature lacks a return type, and the datatype \texttt{array_string} is not recognized by gretl - \texttt{strings} should be used instead. Furthermore, \texttt{array_length()} is not a built-in function in gretl, as the LLM notes. Additionally, the initialization of the string array \texttt{new_array} is incorrect since the array's dimension is not set. As a result, accessing a specific entry in the object (such as \texttt{new_array[i]} within the loop-block) is not possible. Lastly, the string concatenation will not work.

Despite these issues, the proposed function serves as a useful skeleton for developers who want to structure a function that accomplishes the task, given the docstring written in plain language. The LLM provides a highly detailed description of the individual steps, which can be particularly beneficial for fixing the bugs.

\begin{lstlisting}[caption={Code created for prompt in Listing \ref{code:p_write_add_prefix_to_any}. Syntactical error coloured in red.}, label={code:a_write_add_prefix_to_any}]
function @\color{red}<RETURN TYPE IS MISSING>@ array_add_prefix(@\color{red}array_string@ targets, string prefix)
 if @\color{red}array_length@(targets) == 0
 	@\color{red}error@ ValueError: The input array cannot be empty."
 endif
 @\color{red}array_string@ new_array
 loop i=1..@\color{red}array_length@(targets)
 	new_array[i] = @\color{red}prefix \$ targets[i]@
 endloop
 return new_array
end function

This function takes two arguments:

targets: an array of strings to modify
prefix: the prefix to add to each element in the array
It first checks if the input array is empty, and if so, throws a ValueError.

Then, it initializes a new array new_array to store the modified strings, and loops through each element in targets. For each element, it concatenates the prefix with the element using the $ operator, and adds the resulting string to new_array.

Finally, it returns the modified array new_array.

Note that I assumed array_length is a built-in function in gretl to get the length of an array. If this is not the case, you may need to define it yourself or use a different method to check for an empty array.
\end{lstlisting}

Listing \ref{code:p_write_rmse} shows how we asked the LLM to generate a function that calculates the root mean squared forecast error, based on a simple docstring. The resulting function, as shown in Listing \ref{code:a_write_rmse}, has the correct signature and is well-structured with only one syntactical error. Specifically, the gretl function \texttt{sumsq()} is not available, and we need to replace it with \texttt{$e. \hat{}\; 2$} to correctly calculate the sum of squared errors. Although the LLM output requires some minor modifications, it provides almost working and correct code. Additionally, the LLM provides a detailed description of each line of code which is makes it easy to understand.

\begin{lstlisting}[caption={Prompt to write a gretl function for computing the root mean squared forecast error.}, label={code:p_write_rmse}]
Write a gretl function using the following docstring. The docstring follows in triple backtics.
```
Compute the root mean squared forecast error.

:param y: Series of realizations
:type y: series
:param f: Series of forecasts
:type f: series
:rtype: scalar
```
\end{lstlisting}

\begin{lstlisting}[caption={Code written for computing the root mean squared forecast error. Syntactical error coloured in red.}, label={code:a_write_rmse}]
```
Here's a gretl function that computes the root mean squared forecast error:
```
function scalar rmse_forecast_error(series y, series f)
   matrix e = y - f
   scalar n = nelem(e)
   scalar mse = 1/n * @\color{red}sumsq(e)@
   return @\color{red}sqrt(mse)@
end function
```

The function takes in two series: `y` and `f`, which represent the realizations and forecasts, respectively. It then computes the error for each forecast by subtracting the forecast from the corresponding realization, and stores these errors in a matrix `e`.

Next, it computes the mean squared error (MSE) by dividing the sum of squared errors (`sumsq(e)`) by the sample size (`n`). Finally, it returns the square root of the MSE, which is the root mean squared forecast error.

Note that we used the `nelem()` function to calculate the sample size. This function returns the number of elements in a matrix or a series. We also used the `sumsq()` function to calculate the sum of squared errors. This function takes a matrix or a series as an argument and returns the sum of squared values.
\end{lstlisting}

In another example, we requested the LLM to write a function that calculates the Fibonacci sequence, which is presented in Listing \ref{code:p_write_fibonacci}.

\begin{lstlisting}[caption={Prompt to write a gretl function for computing the Fibonacci sequence.}, label={code:p_write_fibonacci}]
Write gretl code given the following docstring in triple backtics.
```
Fibonacci sequence is a sequence in which each number is the sum of the two preceding ones. The sequence starts from 0 and 1.

:param steps: Length of the sequence
:type steps: integer of the length of the sequence
:return: Fibonacci number
:rtype: integer
:raises: Error if the parameter steps is not positive.
```
\end{lstlisting}

\begin{lstlisting}[caption={Code written for computing the Fibonacci sequence. Syntactical error coloured in red.}, label={code:a_write_fibonacci}]
```
function @\color{red}integer@ fibonacci(@\color{red}integer@ steps)
	if (steps <= 0)
		@\color{red}errprintf@("Error: steps parameter must be a positive integer.\n")
		return
	endif

	if (steps == 1)
		return 0
	endif

	if (steps == 2)
		return 1
	endif

	series fib = zeros(steps)
	fib[1] = 0
	fib[2] = 1

	loop i = 3..steps
		fib[i] = fib[i-1] + fib[i-2]
	endloop

	return fib[steps]
end function
```

This function takes an integer parameter `steps` and returns the `steps`th number in the Fibonacci sequence. If the `steps` parameter is not positive, the function raises an error and returns `null`. The function first checks if `steps` is less than or equal to zero. If so, the function raises an error and returns `null`. Next, the function checks if `steps` equals 1 or 2 and returns the first or second number of the Fibonacci sequence respectively. For `steps > 2`, the function calculates each of the Fibonacci numbers up to `steps` using a loop that starts at 3 and continues until the `steps`th number is reached. It then returns the `steps`th number of the sequence.
\end{lstlisting}

The function name and variable names used in the code are clear and descriptive, and the structure of the code looks satisfactory at first glance. However, there are some minor errors in the code that need to be addressed. First, the type of value returned should be \texttt{scalar} instead of \texttt{integer} when using gretl. Additionally, the command \texttt{errprintf()} does not exist and should be replaced with the \texttt{funcerr} command. Moreover, the object holding the Fibonacci series should be initialized as a vector using the command \texttt{matrix fib = zeros(steps)}, instead of being initialized as a type series.
Once these minor corrections have been made, the proposed code can be executed in gretl to return accurate results.

Our docstring-to-code exercises demonstrate that LLMs can provide significant support for generating gretl code. Specifically, they can serve at the very least as a starting point by providing the general structure of a program. However, as expected, LLMs may not be fully aware of gretl functions, datatypes, and syntax, so their generated code may not be fully functional or may require further editing. Nonetheless, the ability of LLMs to generate accurate and helpful code to some extent highlights their potential as a valuable tool for developers and researchers in the field of programming. Also, it substantially lowers the entry barrier as the level of basic programming skills required decrease (low-level programming).

\subsection{Code explanation and summarization}
Research into using LLMs for summarizing code is currently active \parencite{Ahmed.Devanbu2022}. Programmers may encounter functions that are difficult to understand due to poor documentation, convoluted programming, or specialized mathematical concepts that require domain knowledge. Recent studies by \textcite{Tian.etal2023} have evaluated ChatGPT and found that it can produce useful explanations of code to a high degree. The authors note that even for incorrect code, ChatGPT can reason the intended purpose of the code, although its explanation may not accurately describe the current code.

To study the complexity of the \texttt{kernel_loop()} function \parencite{Lucchetti.Valentini2023}, we chose to use it as an example. We ensured that our prompt-session had no previous knowledge of this function by operating within a new session. Our goal was to evaluate whether the LLM could accurately explain the intention of the code. Listing \ref{code:p_explain_kernel_loop} shows the prompt requesting an explanation of the function.

\begin{lstlisting}[caption={Prompt to explain the function \texttt{kernel_loop()}}, label={code:p_explain_kernel_loop}]
What does the following gretl function do? Please explain. The function is provided in triple backtics:
```
function matrix kernel_loop (const matrix y,
								const matrix X,
								const matrix Z,
								scalar h,
								scalar type[0:4:1],
								matrix optparm[null])
	scalar q = cols(y)
	scalar k = cols(X)
	scalar T = rows(y)
	matrix ret = mshape(NA,T,q*k)
	W = kernelmat(T, type, h, optparm)
	ini = T + 1
	fin = 2*T
	loop t = 1 .. T
		matrix PX = W[ini:fin] .* Z
		matrix num = PX'y
		matrix den = PX'X
		ret[t,] = vec(den\num)'
		ini--
		fin--
	endloop
	return ret
end function
```
\end{lstlisting}

The detailed answer provided by the LLM for the \texttt{kernel_loop()} function is presented in Listing \ref{code:a_explain_kernel_loop}. The LLM correctly identifies the function signature and provides a detailed description of the initialized variables. The most noteworthy aspect of the LLM's output is its correct identification of the matrix \texttt{W} as a weight matrix, indicating that the \texttt{kernelmat()} function returns a vector of weights. The LLM also correctly identifies the element-wise multiplication of "the subset of the \texttt{W} matrix", demonstrating its ability to recognize the gretl syntax \texttt{.*} for element-wise multiplication. Moreover, the LLM's overall statement that "this function computes some sort of weighted regression using the 'y', 'X', and 'Z' matrices, where the weights are determined by the 'W' matrix calculated using the 'kernelmat' function" is accurate and complete.

In another prompt session (not shown here), the LLM provided even more precise information, stating that "the function appears to implement a kernel-based estimation procedure. Specifically, it computes a regression of 'y' on 'X' using instrumental variables generated from 'Z' based on a kernel weighting scheme. The resulting estimates are then used to compute regression coefficients and store them in the 'ret' matrix." This ability of the LLM to recognize the function's use of instrumental variables is particularly significant from an econometric standpoint.

\begin{lstlisting}[caption={Answer explaining the function \texttt{kernel_loop()}}, label={code:a_explain_kernel_loop}]
The function is called `kernel_loop` and appears to be a function defined in the gretl programming language. Here is a brief explanation of what the function does:

The `kernel_loop` function takes several arguments, including `y`, `X`, `Z`, `h`, `type`, and `optparm`. These arguments appear to be matrices and scalars that will be used in the computation.

The function first creates a new matrix `ret` using the `mshape` function with parameters `NA`, `T`, and `q*k`. This will create a matrix with `T` rows and `q*k` columns, where each element is initialized to `NA`.

The function then creates a new matrix `W` using the `kernelmat` function with parameters `T`, `type`, `h`, and `optparm`. This will create a matrix that will be used as a weight matrix.

Next, the function initializes the variables `ini` and `fin` to values `T+1` and `2*T`, respectively. This is used to keep track of the elements in the `W` matrix that will be used in the computation.

The function then enters a loop that iterates through each row of the `y` matrix. For each row, the function first constructs a new matrix `PX` by element-wise multiplying the subset of the `W` matrix specified by the `ini` and `fin` variables with the `Z` matrix. The function then performs some matrix operations involving `PX`, `y`, and `X`, and stores the result in the corresponding row of the `ret` matrix.

After each iteration, the `ini` and `fin` variables are decremented, to select a different subset of the `W` matrix for the next iteration.

Finally, the function returns the `ret` matrix.

Overall, it appears that this function computes some sort of weighted regression using the `y`, `X`, and `Z` matrices, where the weights are determined by the `W` matrix calculated using the `kernelmat` function. The exact nature of the computation will depend on the values of the input variables and the implementation details of the various matrix operations used in the function.
\end{lstlisting}

Next, we provide a second example that uses a function from the SVAR package (v1.97) developed by \textcite{Lucchetti.Schreiber2022}. In Listing \ref{code:p_explain_max_eval}, we present a prompt where we ask the LLM to describe the function \texttt{max_eval()}. Unfortunately, the function's name is ambiguous and lacks specificity, making it difficult to infer its purpose. The use of the term "eval" could signify different things, including "evaluation," "eigenvalue," or other possibilities. Moreover, the function lacks a docstring and any explanatory comments, further compounding the ambiguity. However, we note that the function utilizes several built-in functions from gretl, such as \texttt{eigen()}, to compute the eigenvalues of a square matrix.

\begin{lstlisting}[caption={Prompt to explain the function \texttt{max_eval()}}, label={code:p_explain_max_eval}]
What does the following gretl function do? Explain. The function is in triple backticks.
```
function scalar max_eval (const matrix A)
	n = rows(A)
	p = cols(A) / n
	matrix compan = p==1 ? A : A | (I(n*(p-1)) ~ zeros(n*(p-1), n))
	matrix lambda = eigen(compan)
	# (lambda will be complex now)
	scalar maxmod = maxc(abs(lambda))
	return maxmod
end function
```
\end{lstlisting}

To clarify the function's purpose, we turn to the LLM, which provides additional information about the function in Listing \ref{code:a_explain_max_eval}.

\begin{lstlisting}[caption={Answer explaining the function \texttt{max_eval()}}, label={code:a_explain_max_eval}]
The provided gretl function `max_eval` computes the maximum modulus of the eigenvalues of a square matrix `A`. Here are the steps that the function follows:

The function takes a matrix `A` as an input and extracts the number of rows `n` and number of blocks `p` from `A`. It then constructs a companion matrix `compan` based on `A`. If `p==1`, `compan` is just equal to `A`, otherwise, `compan` concatenates `A` with an identity matrix of size `n(p-1)` on the lower block-diagonal.

The function then computes the eigenvalues of `compan` matrix using the `eigen` function available in gretl.

Finally, it computes the maximum modulus (absolute value) of the complex eigenvalues obtained from the `eigen` function and returns it as the output.

In summary, the function can be used to determine the maximum absolute value of the eigenvalues of a square matrix in gretl. This information can be useful in a variety of applications, such as system simulation or control theory.
\end{lstlisting}

According to the LLM's description, the function \texttt{max_eval}() computes the maximum modulus of a square matrix's eigenvalues. However, we note that the function can also process rectangular matrices, despite the LLM's initial claim. Specifically, \texttt{max_eval}() constructs a companion matrix, named \texttt{compan}, that embeds the input matrix \texttt{A} in a larger square matrix. The LLM's explanation of the function is informative and emphasizes its potential usefulness in system simulation.

Overall, these advanced examples clearly demonstrate the effectiveness of using an LLM to explain econometric functions.

\subsection{Suggestion of variable and function names}
Proper naming of functions and variables is critical for code readability and maintainability. Descriptive and self-explanatory names help to reduce the cognitive load of understanding code and eliminate the need for extensive comments. However, finding good names can be challenging for programmers who would rather focus on code implementation than name selection. In this exercise, we explore whether an LLM can help generate suitable names for functions and variables based on the code body.

To illustrate the process, we use the \texttt{kernel_loop()} function as an example  \parencite{Lucchetti.Valentini2023}. We start by opening a new session to clear the LLM's memory and rename the function to \texttt{foo()} to avoid giving too much away about its purpose. The prompt is presented in Listing \ref{code:p_funcname_foo}, and the output is displayed in Listing \ref{code:a_funcname_foo}. The LLM provides suggestions for function names along with a brief explanation of their meanings.

\begin{lstlisting}[caption={Prompt to suggest function names for function \texttt{foo()}}, label={code:p_funcname_foo}]
Suggest three descriptive function names for the following gretl function. The gretl function is provided in triple backtics:
```
function matrix foo (const matrix y,
								const matrix X,
								const matrix Z,
								scalar h,
								scalar type[0:4:1],
								matrix optparm[null])
	scalar q = cols(y)
	scalar k = cols(X)
	scalar T = rows(y)
	matrix ret = mshape(NA,T,q*k)
	W = kernelmat(T, type, h, optparm)
	ini = T + 1
	fin = 2*T
	loop t = 1 .. T
		matrix PX = W[ini:fin] .* Z
		matrix num = PX'y
		matrix den = PX'X
		ret[t,] = vec(den\num)'
		ini--
		fin--
	endloop
	return ret
end function
```
\end{lstlisting}

The first suggestion, \texttt{kernel_regression}, is descriptive, accurate, and reasonably concise. However, the second suggestion, \texttt{local_linear_regression}, may be ambiguous as to what "local" entails. Regarding the third suggestion, the naming of "h" is not precise enough to be informative. Additionally, the suggested name "estimation_regression" is too general and fails to specify the regression type.

Overall, this exercise shows that an LLM could provide valuable insights into suitable function and variable names, thus enhancing code readability and maintainability for programmers.

\begin{lstlisting}[caption={List of suggested function names for function \texttt{foo()}}, label={code:a_funcname_foo}]
Here are three possible descriptive function names for the provided gretl function:
1. "kernel_regression": This name reflects the fact that the function is performing a kernel regression, which is a type of non-parametric regression that estimates the relationship between variables using a weighted average of neighboring data points.
2. "local_linear_regression": This name also captures the essence of the function, which is to estimate a regression relationship using local (as opposed to global) linear models.
3. "h_estimation_regression": This name emphasizes the fact that the function is using a bandwith parameter h to determine the range of nearby data points that are used in the regression estimation.
\end{lstlisting}

Another example highlights how an LLM can assist in improving variable names. We again employ a function from the "SVAR" gretl package \parencite{Lucchetti.Schreiber2022} called \texttt{CheckNormalizeRd()}. We keep the original comment block at the top of the code and instruct the LLM to limit variable names to a maximum of 10 characters (see Listing \ref{code:p_funcname_CheckNormalizeRd}).

\begin{lstlisting}[caption={Prompt to suggest variable names for function \texttt{CheckNormalizeRd()}}, label={code:p_funcname_CheckNormalizeRd}]
Please suggest improved variable names for the following gretl function in triple backticks. Restrict the length of variable names to 10 characters.
```
function scalar CheckNormalizeRd (matrix *R,
									matrix *d)
	/*
		Checks that
		(1) the constraints are consistent
		(2) the constraints are non-contradictory (??? non-redundant ? - Sven)

		if (1) fails, an error message is printed and R and d are replaced by
		empty matrices; if (2) fails, redundant rows in R and d are dropped.
	*/

	p = rows(R)
	r = rank(R)
	ret = 0

	if r < p
		matrix Rd = R ~ d
		if r < rank(Rd)
			R = {}
			d = {}
			ret = 1

		else
			matrix RR
			matrix QQ = qrdecomp(Rd', &RR)
			matrix e = abs(diag(RR)) .> $macheps
			QQ = selifc(QQ, e')
			RR = selifr(selifc(RR, e'), e)
			Rd = QQ * RR
			R = Rd[1:rows(Rd)-1,]'
			d = Rd[rows(Rd),]'
			ret = 2
		endif
	endif

	return ret
end function
```
\end{lstlisting}

Additionally, we found that the LLM was able to improve the overall readability and ease of comprehension of the function by making the variable names more intuitive and meaningful. The resulting variable names were succinct and clear, providing a significant improvement over the previous, less informative names. Importantly, despite the changes to the variable names, the function itself remained fully functional and accurate. The abbreviations "cons", "Mat," and "Val" correspond to "constrained," "Matrix," and "Value," respectively. This example ultimately demonstrated the utility of the LLM in assisting coders with generating clear and concise variable names that facilitate the understanding of code.

\begin{lstlisting}[caption={Refactored variable names of function \texttt{CheckNormalizeRd()}}, label={code:a_funcname_CheckNormalizeRd}]
Based on the length restriction of 10 characters, here are some suggestions for improved variable names in the `CheckNormalizeRd` function:

- `consMat` instead of `R`
- `consVal` instead of `d`
- `numCons` instead of `p`
- `rankCons` instead of `r`
- `cleanMat` instead of `Rd`
- `decompMat` instead of `RR`
- `selector` instead of `e`
- `cleanedMat` instead of `QQ`

With these name changes, the function becomes:

```
function scalar CheckNormalizeRd (matrix *consMat, matrix *consVal)
	/*
		Checks that
		(1) the constraints are consistent
		(2) the constraints are non-contradictory (??? non-redundant ? - Sven)
		if (1) fails, an error message is printed and consMat and consVal
		are replaced by empty matrices; if (2) fails, redundant rows
		in consMat and consVal are dropped.
	*/

	numCons = rows(consMat)
	rankCons = rank(consMat)
	ret = 0

	if rankCons < numCons
		matrix cleanMat = consMat ~ consVal
		if rankCons < rank(cleanMat)
			consMat = {}
			consVal = {}
			ret = 1
		else
			matrix decompMat
			matrix selector = abs(diag(qrdecomp(cleanMat', &decompMat))) .> $macheps
			matrix cleanedMat = selifc(qrdecomp(cleanMat', &decompMat), selector')
			decompMat = selifr(selifc(decompMat, selector'), selector)
			cleanMat = cleanedMat * decompMat
			consMat = cleanMat[1:rows(cleanMat)-1,]'
			consVal = cleanMat[rows(cleanMat),]'
			ret = 2
		endif
	endif

	return ret
end function
```
\end{lstlisting}

\subsection{Improving code}
Refactoring is a software development technique used to enhance the quality of existing code without impacting its functionality. This process entails modifying the code's internal structure to make it more efficient, maintainable, and readable while removing duplicate code and ensuring compliance with coding standards. Refactoring has numerous benefits for software development projects, including decreased development time and cost, enhanced reliability and performance, and improved codebase maintainability. Several studies have underscored the importance of refactoring in software development projects, including \parencite{Mens.Tourwe2004}.

To illustrate the concept, we utilized the original \texttt{CheckNormalizeRd()} function  from the package written by \textcite{Lucchetti.Schreiber2022}, and instructed the LLM to refactor it according to the "Clean Code" principle (see Listing \ref{code:p_refactor_CheckNormalizeRd}).\footnote{As a note, this approach promotes the development of simple, readable, efficient, and maintainable code, coined by software expert Robert C. Martin, a.k.a. "Uncle Bob."}

\begin{lstlisting}[caption={Prompt to improve function \texttt{CheckNormalizeRd()}}, label={code:p_refactor_CheckNormalizeRd}]
Suggest improvements for the following gretl function in triple backticks. Make use of the "Clean Code" approach.
```
function scalar CheckNormalizeRd (matrix *R,
									matrix *d)
	/*
		Checks that
		(1) the constraints are consistent
		(2) the constraints are non-contradictory (??? non-redundant ? - Sven)

		if (1) fails, an error message is printed and R and d are replaced by
		empty matrices; if (2) fails, redundant rows in R and d are dropped.
	*/

	p = rows(R)
	r = rank(R)
	ret = 0

	if r < p
		matrix Rd = R ~ d
		if r < rank(Rd)
			R = {}
			d = {}
			ret = 1

		else
			matrix RR
			matrix QQ = qrdecomp(Rd', &RR)
			matrix e = abs(diag(RR)) .> $macheps
			QQ = selifc(QQ, e')
			RR = selifr(selifc(RR, e'), e)
			Rd = QQ * RR
			R = Rd[1:rows(Rd)-1,]'
			d = Rd[rows(Rd),]'
			ret = 2
		endif
	endif

	return ret
end function
```
\end{lstlisting}

Listing \ref{code:a_refactor_CheckNormalizeRd} shows the proposed refactoring of the code body. Apart from some formatting issues, the LLM attempts to simplify the nested if-else block. While in the original code, the \texttt{ret=0} is returned if the \texttt{rank(r) $\geq$ rows(R)} this was not present in the refactored code resulting in a serious bug. We do not further evaluate the output here. However, this shows that one has to be cautious when using LLMs for programming.

\begin{lstlisting}[caption={Refactored function \texttt{CheckNormalizeRd()}}, label={code:a_refactor_CheckNormalizeRd}]
	Here are some suggested improvements for the `CheckNormalizeRd` function in triple backticks, using the "Clean Code" approach:

	```
	function scalar check_normalize_rd(matrix *R, matrix *d)
		/*
			Checks that the constraints are:
			(1) consistent and
			(2) non-contradictory.
			If (1) fails, an error message is printed and R and d are
			replaced by empty matrices; if (2) fails, redundant rows
			in R and d are dropped.
		*/

		num_constraints = rows(R)
		rank_R = rank(R)

		if rank_R < num_constraints 						# check for inconsistency
			print "The constraints are inconsistent."
			R = {}
			d = {}
			return 1

		elif rank_R == num_constraints					# check for redundancy
			return 0

		else
			matrix Rd = R ~ d
			matrix RR
			matrix QQ = qrdecomp(Rd', &RR)
			matrix e = abs(diag(RR)) .> $macheps
			QQ = selifc(QQ, e')
			RR = selifr(selifc(RR, e'), e)
			Rd = QQ * RR
			R = Rd[1:rows(Rd)-1,]'
			d = Rd[rows(Rd),]'
			return 2
		endif
	end function
	```

	1. Function names should be in lowercase, so `CheckNormalizeRd` should be changed to `check_normalize_rd`.

	2. Add a brief description of what the function does in the comments.

	3. Use descriptive variable names instead of single character variables. For example, `p` and `r` should be changed to `num_constraints` and `rank_R`, respectively.

	4. Simplify the conditionals for consistency and redundancy checks.

	5. Use early returns instead of nested conditionals to reduce indentation.

	6. Refactor the code to avoid duplicating codes within the `if` and `else` cases.

	7. Add spaces between the operators and operands for better readability.

	Applying these changes can make the function easier to read, understand, and modify.
\end{lstlisting}

We present another example in Listing \ref{code:p_refactor_max_eval}. The function \texttt{max_eval()}, sourced from the SVAR package (v1.97), is examined. This function utilizes linear algebra, specifically calling gretl built-in functions and a ternary operator abbreviation for an if-else block. Despite the absence of descriptive comments or docstrings, the function is complex and necessitates an understanding of linear algebra.

\begin{lstlisting}[caption={Prompt to improve function \texttt{max_eval()}}, label={code:p_refactor_max_eval}]
Suggest improvements for the following gretl function in triple backticks. Make use of the "Clean Code" approach.
```
function scalar max_eval (const matrix A)
	n = rows(A)
	p = cols(A) / n
	matrix compan = p==1 ? A : A | (I(n*(p-1)) ~ zeros(n*(p-1), n))
	matrix lambda = eigen(compan)
	# (lambda will be complex now)
	scalar maxmod = maxc(abs(lambda))
	return maxmod
end function
```
\end{lstlisting}

The output produced is intriguing and displayed in Listing \ref{code:a_refactor_max_eval}. The LLM comprehends the function's purpose, calculating the eigenvalue modulus of a matrix, and identifies two possible scenarios: that the input is either a square or block matrix. It enhances the function's readability by supplementing a docstring, adjusting variable names, and implementing an if-else block while avoiding ternary operators. Additionally, the code is simplified so that each line performs a single operation, in accordance with the "Clean Code" philosophy.

\begin{lstlisting}[caption={Answer of improved function \texttt{max_eval()}}, label={code:a_refactor_max_eval}]
Here are a few suggestions for improving the `max_eval` function using the "Clean Code" approach:

```
function scalar max_eval(const matrix A)
	/*
	Computes the maximum modulus of all eigenvalues of the matrix A.
	If A has multiple columns, it is assumed that it is arranged in a
	stacked block matrix format with n x n submatrices, and the function
	computes the maximum eigenvalue modulus of the matrix A as if it were
	block diagonal. If A has only one column, it is treated as a square
	matrix of size n x n.
	*/
	num_rows = rows(A)
	num_cols = cols(A)
	is_block_matrix = num_cols > num_rows

	if is_block_matrix
		num_blocks = num_cols / num_rows
		identity_matrix = I(num_rows * (num_blocks - 1))
		zeros_matrix = zeros(num_rows * (num_blocks - 1), num_rows)
		stacked_matrix = A | (identity_matrix ~ zeros_matrix)
		eigenvalues = eigen(stacked_matrix)
	else
		eigenvalues = eigen(A)
	endif

	max_eigenvalue_modulus = maxc(abs(eigenvalues))

	return max_eigenvalue_modulus
end function
```

1. Add a brief description of what the function does in the comments.

2. Use descriptive variable names instead of single character variables. For example, `n`, `p`, and `compan` should be changed to `num_rows`, `num_blocks`, and `stacked_matrix`, respectively.

3. Add whitespace around operators to improve readability.

4. Add conditional logic to handle the case where `A` has multiple columns. In this case, it is assumed that the columns represent stacked block matrices, where each block is a square matrix of size n x n.

5. Instead of using a ternary operator to define `compan`, use an `if` and an `else` clause to make the code more readable.

6. Refactor the code to reduce indentation by using an early return.

7. Add more complete documentation to describe how the function works, what type of input it takes, and what output it produces.

8. Use consistent and meaningful variable names, such as `is_block_matrix` and `max_eigenvalue_modulus`.

By applying these changes, the resulting function becomes easier to understand and modify.
\end{lstlisting}

Overall, these examples demonstrate that LLMs may be used to improve code quality through refactoring. However, the output must be scrutinized.

\subsection{Write unit tests}
Unit testing is standard in the modern programming paradigm for detecting bugs in units of a program. However, writing high-quality unit tests can be time-consuming. \textcite{Yuan.etal2023} study the performance of ChatGPT to  generate unit tests. They find that the LLM suffers from correctness issues including diverse compilation errors and execution failures. Still, the authors interpret the results as promising for automatically writing unit tests.

To evaluate the LLM's ability to write unit tests for gretl, we employed the refactored \texttt{max_eval()} function from the previous subsection. In Listing \ref{code:p_unittest_max_eval}, we prompted the LLM to write a unit test for this function. The unit test should contain an artificial dataset and compare the actual output with an expected result for both the square and block matrix cases.

\begin{lstlisting}[caption={Prompt to write a unit test for function \texttt{max_eval()}}, label={code:p_unittest_max_eval}]
Please write a unit test using the gretl syntax for the following gretl function in triple backticks.
```
function scalar max_eval(const matrix A)
	/*
	Computes the maximum modulus of all eigenvalues of the matrix A.
	If A has multiple columns, it is assumed that it is arranged in a
	stacked block matrix format with n x n submatrices, and the function
	computes the maximum eigenvalue modulus of the matrix A as if it were
	block diagonal. If A has only one column, it is treated as a square
	matrix of size n x n.
	*/
	num_rows = rows(A)
	num_cols = cols(A)
	is_block_matrix = num_cols > num_rows

	if is_block_matrix
		num_blocks = num_cols / num_rows
		identity_matrix = I(num_rows * (num_blocks - 1))
		zeros_matrix = zeros(num_rows * (num_blocks - 1), num_rows)
		stacked_matrix = A | (identity_matrix ~ zeros_matrix)
		eigenvalues = eigen(stacked_matrix)
	else
		eigenvalues = eigen(A)
	endif

	max_eigenvalue_modulus = maxc(abs(eigenvalues))

	return max_eigenvalue_modulus
end function
```
\end{lstlisting}

Although the output in Listing \ref{code:a_unittest_max_eval} appeared reasonable at first glance, closer inspection revealed several issues. For the square matrix case, the LLM incorrectly created the matrix using \texttt{matrix A = { 1, 2; 3, 4 }} instead of gretl syntax.\footnote{In gretl, creating a matrix is done by \texttt{matrix A = \{ 1, 2; 3, 4 \}}.} Additionally, its expected maximum eigenvalue of 5.0 was incorrect, as the correct value was 5.372. Furthermore, the LLM was not familiar with gretl's built-in \texttt{assert()} function for invoking assertions. Thus, the unit test would fail even though the function returns the correct value.

For the stacked block matrix case, the LLM provided a poor example of a unit test. It again failed to properly create a matrix using gretl syntax, and the proposed matrix \texttt{B} was square rather than block, thus failing to test the intended scenario. The expected maximum eigenvalue of 13 was also incorrect, as the correct value was 13.15.

\begin{lstlisting}[caption={Proposed unit test for function \texttt{max_eval()}. Syntactical error coloured in red.}, label={code:a_unittest_max_eval}]
	Here's an example unit test for the `max_eval` function in Gretl syntax:

	```
	# Test for a square matrix
	matrix A = @\color{red}\{ \{1, 2\}, \{3, 4\} \}@
	scalar expected_result = 5    # max eigenvalue modulus of A is 5
	scalar test_result = max_eval(A)
	printf "Test 1 - Square Matrix: "
	if test_result == expected_result
		print "Passed"
	else
		print "Failed"
	endif

	# Test for a stacked block matrix
	matrix B = @\color{red}\{ \{1, 2, 0, 0\}, \{3, 4, 0, 0\}, \{0, 0, 5, 6\}, \{0, 0, 7, 8\} \}@
	scalar expected_result = 13  # max eigenvalue modulus of B is 13
	scalar test_result = max_eval(B)
	printf "Test 2 - Stacked Block Matrix: "
	if test_result == expected_result
		print "Passed"
	else
		print "Failed"
	endif
	```

	In this example, we run two tests. The first test uses a 2 x 2 square matrix and verifies that the `max_eval` function returns the expected result of 5. The second test uses a stacked 2 x 2 block matrix and verifies that the `max_eval` function returns the expected result of 13. We use `printf` to print informative messages to the console, and we use `if` statements to check if the function's output matches the expected output.
\end{lstlisting}

Overall, while the LLM's unit test appeared to be well-structured, it struggled to create an artificial dataset using gretl syntax and produced incorrect expected values. Additionally, it failed to test the two distinct cases that the function comprised.

\subsection{A tiny coursework example}
The following prompt describes an exercise for an introductory econometrics course. The student is asked to load a dataset, manipulate a variable, compute summary statistics, and run an ordinary least square (OLS) regression (refer to Listing \ref{code:p_write_script}).

\begin{lstlisting}[caption={Prompt to write a gretl script}, label={code:p_write_script}]
I want a gretl script doing the following tasks:
1. Load the "denmark" dataset.
2. Create the first differences of the series "LRM" and "LRY".
3. Compute summary statistics of the series created at step 2.
4. Run a OLS regression. The endogenous variable is the first difference of "LRM". Regress on a constant and the first difference of "LRY".
5. Store the fitted values as "yhat".
6. Plot the fitted values over time
\end{lstlisting}

The output in Listing \ref{code:a_write_script} is noteworthy, with only two minor errors. The script correctly loads the time-series dataset shipped by gretl and creates the desired variable using the built-in \texttt{diff()} function. The LLM is aware that first differencing results in the loss of the initial observation and attempts to set the sample start to the second observation, which is correct. However, the syntax for the \texttt{smpl} command is incorrect; it should be \texttt{smpl 2 ; } which restricts the sample to start with the second observation.
The \texttt{summary} command is used correctly for computing descriptive statistics, followed by the \texttt{ols} command to run the OLS regression and the call to the internal \texttt{\$yhat} accessor to store fitted values. Lastly, the LLM attempts to call the \texttt{gnuplot} command, but the syntax is incorrect; it should be \texttt{gnuplot yhat -{}-time-series -{}-with-lines -{}-output=display}.

\begin{lstlisting}[caption={Proposed gretl script. Syntactical error coloured in red.}, label={code:a_write_script}]
open denmark.gdt

# create first differences of "LRM" and "LRY"
series dLRM = diff(LRM)
series dLRY = diff(LRY)

# compute summary statistics of the series created at step 2
smpl 2@\color{red}:@ --quiet # this sets the sample from the second observation to the end
summary dLRM
summary dLRY

# run OLS regression and store fitted values as "yhat"
ols dLRM const dLRY --quiet
series yhat = $yhat

# plot fitted values over time
gnuplot yhat @\color{red}vs time@
\end{lstlisting}

Next, we engage in iterative prompt development and send the prompt to the LLM to remove the \texttt{smpl} command.\footnote{In gretl all the following commands automatically deal with missing values such that we do not need to restrict the sample explicitly.} The returned output includes the same script but without the \texttt{smpl} call (not shown).

\begin{lstlisting}
Good job. However, get rid of the "smpl" command which is not needed.
\end{lstlisting}

The returned output consists of the same script just without the call of the \texttt{smpl} command (not shown here).

We then ask the LLM to correct the \texttt{gnuplot} command's syntax to \texttt{gnuplot yhat -{}-with-lines -{}-time-series -{}-output=display}. However, despite the correction, the output for this command remains incorrect (shown below).

\begin{lstlisting}
Also, the syntax calling the "gnuplot" command is not correct. Please fix that. Add the options to:
1. Plot the time-series as a line.
2. Output the plot on the screen.
3. And add the time-series option.
\end{lstlisting}

\begin{lstlisting}
	.
	.
	.
# plot fitted values over time
gnuplot yhat @\color{red}with lines time
\end{lstlisting}

\section{Conclusion}
In conclusion, this study has examined the usage and effectiveness of current LLM models for programming with a low-resource and domain-specific programming language. As an application, we choose gretl and its scripting language hansl. We have employed the LLM provided by \url{www.you.com} which is based on GPT-3.5 when writing this article. The findings show that publicly available LLMs already can be a useful tool for understanding, writing, and improving gretl code, despite the fact that only little gretl code is publicly available. This indicates that current LLMs are able to generalize well to LRPL.

Specifically, the LLM produced useful and descriptive docstrings for gretl functions, translates docstrings back to gretl code and \textit{vice versa}, helped to improve the readability and maintainability of code by suggesting better function and variable names, and provided precise and technical explanations of abstract and poorly documented econometric code. Also, we showed how the LLM helps refactoring gretl code mainly involving linear algebra. However, the LLM was not always successful in improving code and also failed to write a correct unit test. Lastly, we presented a simple exercise for an introductory econometrics course. The written script by the LLM is useful as a starting point for students as the syntactical errors are of minor type. We have shown, that the LLM is expected to correct some of the syntactical errors by means of iterative prompt development.

Future research could build on these findings by exploring ways to fine-tune LLM models for gretl code. It also would be interesting to evaluate whether a modern LLM helps to detect and correct errors in gretl code. Lastly, LLMs may be used to translate code from another language into gretl which is a topic under active research \parencite{Roziere.etal2020}.

Overall, this study provides valuable insights into the potential uses and limitations of LLMs in programming with the low-resource and domain-specific econometric language gretl.

%
%

\newpage


\singlespacing
\printbibliography




\end{document}